# Translating Politeness across Cultures: Case of Hindi and English

**Ritesh Kumar**
Jawaharlal Nehru University
New Delhi, INDIA
riteshkrjnu@gmail.com

**Girish Nath Jha**
Jawaharlal Nehru University
New Delhi, INDIA
girishjha@gmail.com

**ABSTRACT**
In this paper, we present a corpus based study of politeness across two languages-English and Hindi. It studies the politeness in a translated parallel corpus of Hindi and English and sees how politeness in a Hindi text is translated into English. We provide a detailed theoretical background in which the comparison is carried out, followed by a brief description of the translated data within this theoretical model. Since politeness may become one of the major reasons of conflict and misunderstanding, it is a very important phenomenon to be studied and understood cross-culturally, particularly for such purposes as machine translation.

**INTRODUCTION**
Polite (or, politic) behaviour has been defined as "socioculturally determined behaviour directed towards the goal of establishing and/or maintaining in a state of equilibrium the personal relationships between the individuals of a social group". [14] It is to be noted that politeness is 'socioculturally determined', which implies that it differs across different societies and cultures, thereby, being a fertile breeding ground for intercultural misunderstandings and conflicts. And indeed very often we come across comments like Japanese are very polite, Englishmen are very courteous and Americans are very rude.

However such comments, as the politeness studies in recent times have pointed out, are misconceptions and bases on false assumptions. It is now a commonly accepted fact that politeness is a pan-cultural, universal phenomenon and every language, spoken anywhere in the world has ways of expressing politeness. Along with this understanding, another fact that has come to the fore as a result of politeness studies in different cultures is that it is expressed in vastly different ways in different languages.

Despite the differences in the way politeness is expressed across languages, over the last four decades or so, there have been several attempts at formalising, universalising and giving a definite direction to this extremely complex and fascinating aspect of human speech and communication. Of special significance and importance in this regard is the seminal work of Brown and Levinson [2] which tried to give mathematical rigour, precision and straightforwardness to the politeness studies. They proposed a theory based on Goffman's concept of negative and positive face and gave a kind of algorithm for explaining what kind of politeness strategy is being used by speakers in a particular instance. Some of the other well-meaning attempts to formalise and model politeness so that it could be studied cross-linguistically include those by Leech [10] and Lakoff [9]. However these theories could not generate the same kind of response as Brown and Levinson's theory.

In spite of all its claims to universality, the model by Brown and Levinson has been hugely attacked by the likes of Eelen [5] and Ide [8] on the grounds of it not being applicable to the non-European languages. Following this criticism there have been several attempts to defend, extend and modify the theory by O'Driscoll [12], Fukada and Asato [7], Meier [11], Pfister [13] and others. Furthermore there have been some attempts by Culpeper [3, 4] and Bousfield [1] to adapt Brown and Levinson's model to study impoliteness in language.

However these and several other attempts have not proved to be very successful in bringing out some universal aspects of politeness. For example, we can take the case of indirect question as a polite way of requesting something, which was considered quite universal but then one of the studies by Wierzbicka [15] showed that a question would be taken as a genuine question in Slavic languages like Polish and Russian instead of as a request. Esacandell-Vidal [6] in her paper talks about a work in Thai which points out that the effect of a question on a Thai partner would be opposite of a polite request. (S)he will think that you are questioning his/her ability and will become very angry. In such situation it becomes imperative that we understand the differences across cultures while trying to find out the universality, since understanding these differences is very necessary for checking intercultural bias, prejudices and conflicts.

**METHODOLOGY AND DATA**
Since most of the models proposed for politeness have failed to accommodate data from across languages, they are not sufficient and good enough for carrying out a comparative analysis of politeness in two languages. We have proposed a 'structural model of politeness' both for the description of politeness in a language as well as for the comparison of politeness across languages. This model is based on the basic classification of the structure of language. This model takes into account the strategies or maxims of the earlier models of politeness However these strategies and maxims are described in terms of their structure. Depending on the way it is expressed, politeness in any language, in general, operate at three different structural levels. These three basic levels at which politeness strategies operate can be termed as

a) <u>Lexical level</u> – An utterance would be polite at the lexical level when there are some word(s) in the utterance which is making the utterance polite. We can take a look at the following examples:

> (i) **Excuse me**, can I borrow your notes **please**? (English).
>
> (ii) **kəhije** rɑhul sinhɑ **ɟi** . (Hindi)
>
>   Please speak Rahul Sinha ji
>
> (iii) **ɑp** jəhã bɛtʰnɑ **pəsənd̪** kəreŋge? (Hindi)
>
>   Would you like to sit here?

b) <u>Syntactic level</u> – An utterance is polite at the syntactic level when the syntactic structure or type of the sentence (for example, requests, apologies, etc.) makes the utterance polite. The examples (i) and (iii), along with exhibiting lexical level politeness, also shows the syntactic level politeness. In both the cases a request is being done with the help of an interrogative question.

c) <u>Pragmatic/discourse level</u> – The reading of an utterance being polite or impolite at the lexical or syntactic level is the default reading which implies that, if these utterances are given to a native speaker without giving the context in which they are spoken, then they would be judged as polite utterances. This default reading is not decided arbitrarily; rather it is reached upon by the most common and dominant usages of structures like these in a particular kind of context. And that very context assigns that particular kind of utterance the polite or impolite value. But this does not rule out other contexts where the same utterance may be assigned a different value. For example, we can take the examples (ii) and (iii) above. If the sentence in (ii) is spoken by a father to his son over a telephone call because the son has not called for months then the utterance becomes sarcastic and not polite. Similarly if the utterance (iii) is spoken by the teacher to his student because he is not coming to the class for long and then suddenly he is caught in front of the classroom, it becomes a rebuke and not a polite way of asking something. So every (im)polite utterance works at this level; it is within a context that a particular value could be assigned to the utterance. However this level remains hidden or inactive till the values are being assigned in the default context.

As we can see from this discussion, these three levels are not mutually exclusive and they are not completely autonomous. In fact there seems to be some kind of hierarchical relationship among them, with pragmatic/discourse level at the top of the hierarchy and lexical and syntactic level below it.

In the present paper we present a comparative analysis of the politeness in Hindi to English translation at the first two levels – lexical and syntactic. This is mainly because of the nature of the corpus that is used for the study and also of the corpus in general. The context remains in the default form throughout the data that has been considered here. Here this default form is the 'formal context', with both the speaker and the hearer being at equal level and having no acquaintance to each other. Thus it is one adult stranger talking to another adult stranger.

The data for the present study is taken from the parallel translation corpus of 50,000 tagged sentences in 12 Indian languages currently being built by the Special Centre for Sanskrit Studies, Jawaharlal Nehru University, New Delhi and 12 other associate universities under the project titled Indian Language Corpora Initiative (ILCI). The basic dataset is being prepared in Hindi and it is then translated into other Indian languages, including English. For the purposes of translation, the rule of structural equivalence or structural parallelism is given the prime importance. The translated texts are intended to have correspondence at lexical (one word is translated into one word and not multiple words and vice-versa) phrasal (one phrase is translated into one phrase and not more than one and vice-versa) as well as clausal level (one clause being translated into only one clause, not more than one and vice-versa). However in certain cases where it is impossible to follow this because either similar structure is not possible in both languages or similar structure makes the translation incomprehensible or useless, this rule may be broken in favour of a better translation.

This structure of the data where a structural similarity is to be maintained while translating (which is an important requirement of any machine translation effort) is very relevant as well as essential for the purposes of the paper. There is also one apparent limitation of the data as it covers only two domains – health and tourism. But on closer examination it also proves to be beneficial as the exhaustive data is provided in these two domains and the study could be later developed exclusively for these two domains.

**TRANSLATING POLITENESS**

While translating any text from Hindi to English (or for that matter from any source language to target language), politeness will be required to be translated at one or more of the above-mentioned structural levels. Consequently theoretically one of the following situations may arise during translation:

**Situation 1: Translatable polite structures**
In this situation both the structure and the politeness value of the sentence is translated at both the levels. It implies that the lexemes making the sentence polite as well as the sentence structure of the source language (SL) is properly translated into the target language (TL). At the same time the politeness value attached to these structures in the source languages remains intact after translation into the TL. This situation is realised when the structure of the SL is also possible in the TL and at the same time that structure is equally polite in both the languages. Our null hypothesis is that this is a highly unlikely situation and we hope to find very little examples of this situation, if at all, we find some.

**Situation 2: Mistranslated polite structures**
In this situation only the structure of the SL is translated into the TL but not the politeness value. This could again be subdivided into two sub-situations – a) mistranslation at the lexical level and b) mistranslation at the syntactic level. This situation arises when the structure of the SL has an equivalent in the TL but this equivalent does not have the same politeness value. This equivalence at the lexical level implies that the given word in SL has an exact one-word equivalent in the TL. Similarly at the syntactic level it implies that the sentence structure of SL is preserved in TL (of course after taking into consideration such variables as the word order distinction in the two languages). Our null hypothesis is that this is a very likely situation in the translations that we are looking at and we hope to quite a lot of examples of this situation.

**Situation 3: Dissimilar structures**
This is an opposite of situation 2. Here the politeness value of a sentence is carried through in the TL, but the structure of the sentence becomes different from the SL. Such kind of translation may arise because of two reasons:

a) The situation is similar to that of situation 2 i.e., the structure is possible in TL but politeness is not carried through. However when the structure of the sentence is modified then the politeness value is preserved. This kind of situation is ruled out in the present study since the corpus that we are using is built on the principle of structural parallelism and so the structures in both the languages have to be similar as far as possible.

b) It is not possible to carry through the structure of the SL into the TL. And so the structure has to be different. However despite this difference in the structure the politeness value is carried through.

Our null hypothesis is that the situation in (b) is again going to be highly unlikely since if the structure is not preserved in transaction then the idea that something like politeness arising out of that structure would be preserved seems quite far-fetched.

**Situation 4: Non-translatable structures**
This situation is similar to the above situation 3 (b). The structure of the SL cannot be translated into the TL. So the structure cannot be preserved. However unlike in the previous situation politeness value is also not carried through in translation. Thus neither the structure nor the politeness value is carried through. Our null hypothesis is that this is also quite a possible situation (although not a very highly likely situation like situation 2).

As a result of the above four situations, following three consequences are expected after translation:

**Consequence 1: Proper Translation**
This consequence arises out of situation 1 and situation 3. Here the politeness value is preserved and the polite sentences in the SL are translated as polite sentences in the TL.

**Consequence 2: Improper Translation**
This consequence arises out of situation 2 and situation 4. Here the politeness value is not preserved and the polite sentence in the SL becomes non-polite (or neutral or politic [16]) in the TL. This consequence is not very dangerous and we have to bear with it any many places. However we should try as much as possible to keep it at bay.

**Consequence 3: Bad Translation**
This consequence again arises out of situation 2 and situation 4. Here also the politeness value is not preserved but it differs from consequence 2 in that polite sentences in SL are translated into impolite sentences in the TL. This is a highly dangerous consequence and needs to be immediately checked and censored. Otherwise it has a great potential to become the source of intercultural friction and conflict.

These three consequences does not cover the whole range of consequences as there could be many steps in between polite and non-polite and non-polite and impolite. Among two sentences one can be said more polite and other less polite, although none could be termed non-polite or impolite. So this scaling will also be fine-tuned as we move along this project.

**LOOKING AT THE CORPUS**
From the little data that we have looked at in the corpus, we have found examples from two of the situations discussed above. We present here a brief snapshot of the kind of data that we have found in different situations. As of now, our hypotheses regarding the situation have turned out to be pretty true.

**Situation 1**
We have yet to come across the data in which polite structures in Hindi are translated into polite structures in English.

**Situation 2**
We have got some examples in this situation. Most of these examples relate to the suggestion:

- bhoʝən ɑsən ke leɡbhəɡ ɑdhe ɡhənʈe pəʃcɑt kərnɑ cɑhije.

One should have meal almost half an hour after the asanas.

- isi prəkɑr gərbhpɑt ki prəvritti wɑli gərbhwətI strijõ me gərbh ki ɑrəmbhikɑwəsthɑ me jɑtrɑ kɑ pəritjɑg kərnɑ lɑbhprəd rəhtɑ hɛ.

Similarly in the women with the tendency of abortion it is beneficial to shun travelling during the initial stages of pregnancy.

**Situation 3**
As per our expectations, we have not found any instance of this situation till now (but even then its possibility is not completely ruled out in the corpus).

**Situation 4**
We have found quite a few examples of this situation. The first example is related to the use of particle 'to' in Hindi, which could not be translated directly into English. In Hindi it is a polite way of asking a question, particularly an uncomfortable question; but in English the use of 'whether' makes the sentence non-polite. A couple of examples are as follows:

- in əŋgõ ko chʊne jɑ dəbɑne pər rogiɲi ko koi kəʂʈ to nəhi hotɑ hɛ.

Whether the patient feels any problem while touching or pressing these body part.

- kəmər ewəm riʈ ki həɖɖi sidhi rəhe.

Back and spinal cord should be straight.

- mɑthɑ dono hɑthõ ke bic ʈikɑ huɑ ho.

Head should be placed in between both the hands.

**THE APPLICATIONS**
The primary application of the paper is expected to be in the field of Machine Translation. The results coming out of this study may be used to formulate rules such that politeness could be properly understood and translated by the machine. The primary aim of these rules would be to check the translation of polite texts into impolite texts.

This study could also be developed as a contrastive study of politeness strategies in Hindi and English, which could be extended to other languages as well. Studies like this could be used for developing a typology of politeness strategies which has long been the aim of politeness theorists.

Moreover it could also be used for pedagogical purposes where the students are systematically told about the contrast between the two languages in such speech acts as requesting, commanding, questioning, apologizing, etc.

**ACKNOWLEDGMENTS**
We thank the whole team of ILCI for providing us the parallel corpus, especially Narayan Chaudhary for translating most of the Hindi texts into English.